\documentclass{INTERSPEECH2023}

\usepackage[dvipsnames]{xcolor}
\usepackage{amsmath,graphicx,dsfont, amssymb,tikz}
\usepackage{array}
\usepackage[font=small]{caption}
\newcolumntype{C}[1]{>{\centering\arraybackslash}p{#1}}

\newcommand{\tikzcircle}[2][black,fill=black]{\tikz[baseline=-0.5ex]\draw[#1,radius=#2] (0,0) circle ;}%

\interspeechcameraready

\title{Evaluating context-invariance in unsupervised speech representations}
\name{Mark Hallap$^{1}$, Emmanuel Dupoux$^{2}$, Ewan Dunbar$^{1,2}$}
\address{
$^1$ University of Toronto\\
$^2$CoML, ENS/CNRS/EHESS/INRIA/PSL Research University}
\email{mark.hallap@utoronto.ca, emmanuel.dupoux@gmail.com, ewan.dunbar@utoronto.ca}

\begin{document}

\maketitle
 
\begin{abstract}
Unsupervised speech representations have taken off with benchmarks demonstrating major progress on semi-supervised speech recognition, speech synthesis, and speech-only language modelling. Inspiration comes from the promise of discovering the phonemes of a language or a similar low-bitrate encoding. However, one of the critical properties of phoneme transcriptions is context-invariance: the phonetic context of a speech sound can have massive influence on the way it is pronounced while text remains stable. This is why tokens of the same word have the same transcriptions---key to language understanding. Current benchmarks do not measure context-stability. We develop a new version of the ZeroSpeech ABX benchmark that does, and apply it to recent self-supervised representations. We show that context-independence of representations is predictive of the stability of word-level representations. We suggest research concentrate on improving context-independence of unsupervised representations.
\end{abstract}
\noindent\textbf{Index Terms}: pre-trained acoustic models, self-supervised speech, unsupervised speech, invariance

\section{Introduction}
\label{sec:intro}

Self-supervised and unsupervised learning of speech representations have their roots in attempts to discover phone- or phoneme-like units, or similarly linguistically relevant representations, either in the interest of working with lower-resource languages  or of cognitive modelling \cite{dupoux2018cognitive}.
These representations have already seen practical success. The use of recent models such as CPC \cite{oord2018cpc}, Wav2vec 2.0 \cite{baevski2020wav2vec}, HuBERT \cite{hsu2021hubert}, or WavLM \cite{chen2022wavlm} for pre-training has been shown to greatly reduce the amount of labelled speech data needed to build a recognizer. But the promise of self-supervised speech representations is much greater. In principle, if we were truly able to discover low-bitrate representations for a language similar to phonemes or letters, higher-level  tasks could be done directly from speech. Currently, however, the performance of self-supervised representations on even mid-level tasks such as word segmentation  lags behind that  of text, to say nothing of higher-level tasks such as language modelling \cite{dunbar2022self}.

What makes text or phoneme representations so fundamental in language processing is their dual nature, linking signifier (form) and signified (content), which otherwise bear no meaningful relation. A language might use any string of sounds to refer to canines, but the transcription of the English word \emph{dog} tells the reader both what was pronounced and (if they know English) what was meant. Text and phonemes sift through all the linguistically insignificant variations in the signal and pinpoint those necessary for understanding.

Critically, phonemes change pronunciation substantially depending on their surrounding phoneme context due to coarticulation and allophony. Yet, current evaluations  of intrinsic   quality for unsupervised and self-supervised representations do not directly measure whether  representations remain the same when the surrounding context changes. In this paper, we propose a novel evaluation based on the ABX phone error rate  which directly assesses context-invariance.

In Experiment 1, we evaluate systems submitted to the 2021 Zero Resource Speech Challenge and demonstrate that effects of context are the most significant source of instability for current models---much greater than lack of invariance to speaker or resistance to less-clean speech. In Experiment 2, we demonstrate that this result is not an artefact of the models' warping of time. Finally, in Experiment 3, we  demonstrate that context-invariance predicts representations' ability to consistently encode tokens of the same word type.

\section{Background}
\label{sec:background}
\label{sec:evals}

In addition to benchmarking on downstream tasks (SUPERB: \cite{yang2021superb}), a large part of self-supervised speech representation evaluation consists of measures of intrinsic quality. Linguistically-motivated measures of intrisic quality have been proposed as part of the ZeroSpeech challenge \cite{dunbar2022self}. Notably, the \textbf{ABX} phone discriminability score \cite{schatz2013} attempts to measure, for a model trained on a given language, how distinctly the model represents  the linguistically relevant sound (phoneme) categories.

The task   is inspired by  human psychophysics and measures discriminability between two sound categories.  
${\Delta}$, the ABX-discriminability of sound category $A$ from category $B$, is defined as the probability that tokens $a,x$  $\in A$ are further apart than token $b \in B$ is from $x$, according to a dissimilarity function $d$. Thus ${\Delta}(A,B)$ is
\begin{multline*}\small
{ \frac{
\sum_{a\in A} \sum_{b\in B} \sum_{x\in A \setminus {a}}(\mathds{1}_{d(a,x)<d(b,x)}  + \frac{1}{2}\mathds{1}_{d(a,x)=d(b,x)})
}{|A|(|A|-1)|B|} }
\label{eq:abx}
\end{multline*}
where $\mathds{1}$ is the indicator function and $|A|$ ($|B|$) the number of tokens in category $A$ ($B$).\footnote{The $d(a,x)=d(b,x)$ clause is particularly important if the representations are symbolic. They need not be, and here for many systems it is continuous embeddings that are used: but the ABX score is compatible with both discrete and continuous representations.} The discriminability score is symmetrized by averaging $\Delta(A,B)$ and $\Delta(B,A)$. 

Evaluating a model for a language begins with the model's encoding for a test corpus. Using the gold alignment, the evaluation splits the encoding  into one sequence of frames for each token of each category. To calculate $d$ for two tokens, dynamic time warping is used to realign them, and frame-level dissimilarities are averaged along the alignment path. Submissions to the ZeroSpeech challenge specify whether to use angular dissimilarity (arccos of the normalized dot product of frame embeddings) or KL divergence to calculate $d$; see \cite{dunbar2022self}. $\Delta$ (discriminability) for all pairs of categories are averaged and subtracted from 1 to obtain an overall ABX error rate.

Importantly, however, in the ZeroSpeech ABX phone evaluation, each token is a triphone: to compare two phoneme categories (for example, /l/ and /r/) we use  triphones where only  the middle phoneme varies in the critical phone contrast---like /flu/--/fru/. The immediate context is thus included in the token, and is constant across $a$, $b$, and $x$. This was originally to avoid relying on very short, one-phone sequences and to avoid demanding that the representations meet the (presumably difficult) requirement of context-independence. The strong scores  on the \emph{within-context} version of the new benchmark, which uses one-phone sequences, demonstrate that the first problem is not an issue with current systems.  

We base our novel evaluation on the \textbf{ABX-LS} benchmark \cite{dunbar2022self} for English, one of the ZeroSpeech ABX benchmark evaluatoins, itself based on LibriSpeech (LS) \cite{panayotov2015librispeech}. This benchmark additionally measures speaker invariance by comparing  a \textit{within-speaker} score, in which  all of the phone triplets belong to the same speaker (e.g., $a=\textrm{flu}_{T_1}$, $b=\textrm{fru}_{T_1}$, $x=\textrm{flu}_{T_1}$) to an  \textit{across-speaker} score, in which $a$ and $b$ are the same speaker and $x$ a different one (e.g., $a=\textrm{flu}_{T_1}$, $b=\textrm{fru}_{T_1}$, $x=\textrm{flu}_{T_2}$). It also  compares the \emph{clean} speech part of the LibriSpeech corpus to the \emph{other} part. The ABX phone discriminability measure has been demonstrated to correlate with the performance of speech representations on a number of other quality measures and downstream tasks \cite{dunbar2022self}.

Another intrinsic quality measure sometimes applied to speech representations is the mean average precision, or \textbf{MAP  on spoken word embeddings} \cite{carlin2011rapid,algayres2020evaluating}.  {MAP} assesses how well spoken word embeddings separate different word types using a method similar (but not equivalent) to the ABX score. All pairs of word tokens in the test corpus are compared using a dissimilarity; then, a precision--recall curve for same/different word type classification is calculated across different dissimilarity threshholds. The {MAP} score is the area under the precision--recall curve. A simple (albeit lossy) way of constructing spoken word embeddings from a self-supervised representation is to simply do mean pooling (time average), which allows for an approximate assessment of the quality of the representation for encoding a spoken lexicon.

We propose an ABX-discriminability measure of how well learned speech representations track context-independent phonemes. The most closely related work is \cite{schatz2017quantitative}, which applies a closely-related ABX- discriminability score to MFCC representations to measure how much different classes of phonemes are acoustically influenced by context, rather than to evaluate learned representations. Additionally, a number of papers have used phoneme classification as an evaluation for the quality of learned speech representations \cite{oord2018cpc,higy2021discrete}. Correct phoneme classification for segments extracted from a representation requires context-independent representations. However,  these classification analyses do not measure context-independence: here, we introduce the critical comparison between a single measure (ABX) in a context-\emph{dependent} versus -\emph{independent} mode to measure context-independence.

\section{Methods}
\label{sec:methods}
\label{sec:neweval}

Starting from \textbf{ABX-LS}, we develop a novel variant to evaluate context-invariance.\footnote{In what follows, we report performance only on the dev subset; results on the test subset are qualitatively the same.} Rather than extracting triphone tokens, we extract phonemes in isolation. In the \emph{within-context} condition, the immediately preceding and following phoneme are held constant across $a$, $b$, and $x$. In the \emph{without-context} condition, there are no constraints on the surrounding context; in general, it varies.   Comparing the two measures invariance to changes in context, i.e., to coarticulation and allophony.

We evaluate all of the self-supervised speech representations submitted to the 2021 ZeroSpeech benchmark \cite{dunbar2022self}, which appear on the current leaderboard.  \textbf{HuBERT}  is an implementation of \cite{hsu2021hubert} described in \cite{nguyen22discrete};  \textbf{CPC}  is an implementation of \cite{oord2018cpc} as described in \cite{nguyen2020zero}. This model was the baseline model for the benchmark, and many submitted systems re-used these baseline features unchanged (taking different approaches to the language modelling component of the benchmark, which does not interest us here). We exclude these submissions. \textbf{S-CPC}   \cite{bhati2021segmental} attempts to push CPC to learn representations of phone segments that are stable across time; \textbf{P\&H VG} is a ``visually grounded'' model \cite{peng2022self} that trains end-to-end on a masked language modelling objective using  spoken picture captions; \textbf{ZR VG} and \textbf{ZR VG-CPC} \cite{alishahi2021zr} are also trained with picture captions, based on \cite{higy_textual_2020}, with the first using MFCCs as input and the second using CPC representations; \textbf{ResDaveNet}, also visually-grounded, is an implementation of \cite{harwath2018jointly}; finally \textbf{CPC+Seg} discovers boundaries on CPC units and applies pooling.


\begin{table*}[ht]
\small
\captionsetup{skip=1pt}
\begin{center}
\begin{tabular}{ccccccccccc}
& \multicolumn{4}{c}{Clean} & &\multicolumn{4}{c}{Other}
\\
& \multicolumn{2}{c}{Within-speaker} &\multicolumn{2}{c}{Across-speaker} &
& \multicolumn{2}{c}{Within-speaker} &\multicolumn{2}{c}{Across-speaker}
\\
&W/in-ctx &W/out-ctx &W/in-ctx &W/out-ctx & &W/in-ctx &W/out-ctx &W/in-ctx &W/out-ctx \\
\hline
HuBERT &\textbf{1.56} &7.26 &\textbf{2.13} &8.04 & &\textbf{3.08} &\textbf{8.64} &\textbf{4.78} &\textbf{10.09} \\
CPC &1.99 &\textbf{7.15} &2.72 &\textbf{7.29} &&4.21 &10.20 &6.65 &11.05 \\
S-CPC &1.99 &{7.17} &2.69 &\textbf{7.29} &&4.21 &10.21 &6.80 &11.08 \\
P\&H VG &2.32 &8.97 &2.80 &10.25 &&4.08 &11.07 &6.46 &13.33 \\
ZR VG-CPC &3.43 &12.26 &4.74 &12.73 &&5.75 &15.23 &9.23 &16.53 \\
ResDaveNet &5.31 &11.58 &6.80 &12.96 &&7.86 &14.29 &11.99 &16.89 \\
ZR VG &5.31 &15.45 &6.89 &16.35 &&7.12 &17.78 &11.41 &19.42 \\
CPC+Seg &5.64 &13.65 &7.24 &13.97 &&10.36 &18.21 &14.77 &19.73 \\\hline
Spectrogram &12.41 &21.13 &19.09 &24.85 &&14.95 &23.32 &23.80 &28.48 \\
\end{tabular}
\end{center}
\caption{ABX context-independence evaluation. ABX error scores (\%). Lower scores are better. Best scores in each column are bolded.}
\label{tab:exp1}
\end{table*}

\section{Experiment 1}
\label{sec:exp1}

\begin{figure}[h]
  \centering
  {\includegraphics[width=9.0cm]{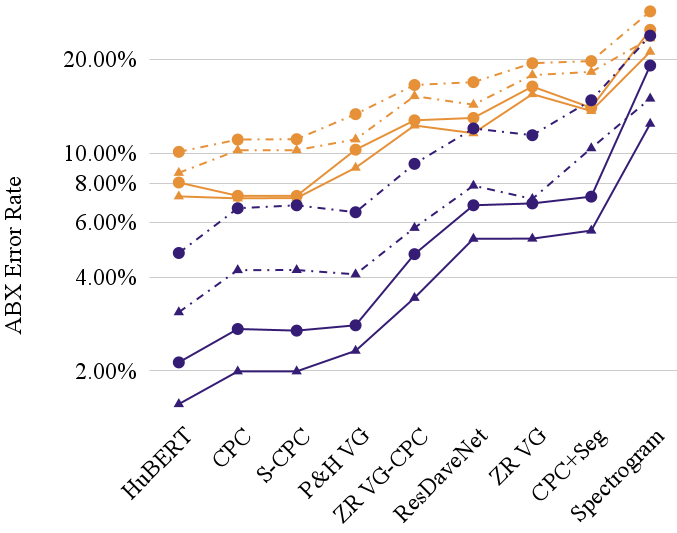}}

\caption{ABX context-independence evaluation scores. Scale is logarithmic. {\color{RoyalPurple}Purple} (darker) lines are \emph{within-context} and {\color{BurntOrange}orange} (lighter) lines are \emph{without-context.} The $\blacktriangle$ points are \emph{within-speaker} while the \tikzcircle{2pt} points are \emph{across-speaker}. Finally the solid lines are \emph{clean} and the dashed lines are \emph{other.}}
\label{fig:exp1}
\end{figure}

Results of applying the novel evaluation to these models are shown in Table \ref{tab:exp1} and Figure \ref{fig:exp1}. The separation between the $\blacktriangle$- and the \tikzcircle{2pt}-pointed lines in Figure \ref{fig:exp1} shows the gap between \emph{within-speaker} and \emph{across-speaker} to vary between a modest penalty and a doubling of the error rate, depending on the model and the evaluation condition. The gap between solid and dashed lines compares the \emph{clean} speech and \emph{other} subsets, and is of a similar magnitude. However, the gap between the darler {\color{RoyalPurple}purple}  \emph{within-context} and the lighter {\color{BurntOrange}orange} \emph{without-context} scores is much greater, tripling or even quadrupling the error rate. This lack of context-invariance disproportionately affects high-performing models such as \textbf{HuBERT} and \textbf{CPC,} suggesting these models' loss encourages them to perform particularly well within context. 

\section{Experiment 2}
\label{sec:exp2}

While Experiment 1 suggests the representations evaluated are not context-invariant, an alternate explanation is  that their time-scales are incompatible with that of the ABX evaluation. The ABX evaluation extracts representations from time-stamps based on gold-standard alignments. Not all model loss functions penalize time warping, and so some representations may preserve the order of phonemes without preserving their exact position in the sequence. If the evaluation is using the ``wrong'' time-alignment for the representation, this would have a disproportionate impact on the without-context condition. Because the context around the phoneme is different, small disagreements in alignment between the representation and the gold transcription will be more important.

To assess this,  we change the way we deal with  time  in the ABX  calculation. Instead of aligning $a$ and $x$ ($b$ and $x$) using DTW and averaging dissimilarities for the aligned frames, we perform pooling to obtain a single representation for $a$, for $b$, and for $x$, weighted using a Hamming window, which peaks at the centre of the sequence; $d$ is  angular dissimilarity or KL divergence. This is less forgiving than DTW, which can compensate for minor errors in the alignment. If the alignment is problematic, ABX scores should worsen when the Hamming window is used. 
In particular, some pairs of representations A--X may be similar to each other but temporally misaligned with each other. Pooling with a filter that peaks in the middle therefore tends to make the inconsistency between A and X worse, particularly if the surrounding context material is not necessarily constant. DTW, on the other hand, can compensate to some degree. 

We perform the experiment  on the \emph{within-speaker, clean} condition.
To  validate the experiment, we evaluate manipulated gold phoneme transcriptions (one-hot, by frame) wherein phoneme boundaries were sometimes ($p=0.5$) shifted right by 4, 6, or 8 frames. We expect scores to be non-zero for the manipulated representations.

\begin{table}[h]
\small
\captionsetup{skip=1pt}
\begin{center}
\begin{tabular}{C{19mm}C{8mm}C{13mm}C{-5mm}C{8mm}C{13mm}}
& \multicolumn{2}{c}{Within-context} &&\multicolumn{2}{c}{Without-context}\\
& DTW &Hamming & &DTW &Hamming \\
\hline
Gold +2 &1.84 &  3.70 & &2.95 & 4.46 \\
Gold +4 &{7.91} &10.05 & &{9.73} &10.23 \\
Gold +6 &{14.67} &16.17&  &{16.26} &16.67 \\
Gold +8 &{21.09} &21.46 & &{21.62} &21.96 \\
Spectrogram &{12.41} &14.05&  &21.13 &{20.34} \\
\hline
HuBERT &{1.56} &1.91 & &{7.26} &7.86 \\
CPC &{1.99} &2.18 & &7.15 &{6.91} \\
S-CPC &{1.99} &2.18 & &7.17  &{6.97} \\
P\&H VG &{2.32} &2.51 & &{8.97} &9.52 \\
ZR VG-CPC &3.43 &{3.42} & &{12.26} &12.59\\
ResDaveNet &{5.31} &6.05 & &{11.58} &12.44 \\
ZR VG &{5.31} &5.48 & &{15.45} &15.71 \\
CPC+Seg &{5.64} &6.18 & &13.65 &{13.20} \\
\hline
\end{tabular}
\end{center}
\caption{ABX scores (\%) comparing DTW (less sensitive to alignment) and Hamming (more sensitive) methods.}
\label{tab:exp2}
\end{table}

Results are shown in Table \ref{tab:exp2}. Frame-shifting the gold transcription indeed makes scores worse and without-context evaluation is more impacted; pooling amplifies the effect.\footnote{Using pooling on the +8 frame-shifting has no clear effect, perhaps because the representation is already so degraded that it only makes coarse-grained contrasts that are somewhat resistant to further perturbation.}
The spectrogram does not appear to be perfectly ``aligned,'' nor do the self-supervised representations (Hamming pooling makes the scores worse): information about phone identity is not all concentrated at the centre of the gold segment.  However, unlike for the manipulated alignments, the difference  ($<$14\% of the original error rate) falls far short of explaining the penalty incurred in the \emph{without-context} condition. We conclude that alignment is at most a small part of the reason for systems' issues with context-invariance.

\section{Experiment 3}
\label{sec:exp3}

\begin{table}[h]
\small
\captionsetup{skip=1pt}
\begin{center}
\begin{tabular}{C{18.2mm}C{5mm}C{16.5mm}C{18.4mm}C{8mm}}
&\textbf{Filter} &$\downarrow$\textbf{ABX w/in} &$\downarrow$\textbf{ABX w/out} &$\uparrow$\textbf{MAP} \\\hline
HuBERT &None & 2.13 & 8.04 &48.25 \\
HuBERT &3    & 2.06 &9.24   &47.63 \\
HuBERT &5    & 2.50 &11.92  &45.86 \\
HuBERT &7    & 3.20 &15.34  &42.90 \\\hline
CPC    &None & 2.72 &7.29 &32.93 \\
CPC    &3    & 2.75 &7.55 &32.73 \\
CPC    &5    & 2.86 &8.25 &32.31 \\
CPC    &7    & 3.05 &9.34 &31.66 \\\hline
S-CPC  &None & 2.69 &7.29 &32.93 \\
S-CPC  &3    & 2.69 &7.56 &32.73 \\
S-CPC  &5    & 2.80 &8.25 &32.31 \\
S-CPC  &7    & 2.98 &9.34 &31.66 \\\hline
P\&H VG &None& 2.80 &10.25 &42.77 \\
P\&H VG &3   & 2.95 &11.63 &41.90 \\
P\&H VG &5   & 3.54 &14.60 &39.91 \\
P\&H VG &7   & 4.61 &18.18 &36.88 \\\hline
ZR VG-CPC&None& 4.74&12.73 &33.55 \\
ZR VG-CPC&3  & 4.85 &12.23 &33.19 \\
ZR VG-CPC&5  & 5.43 &16.63 &32.47 \\
ZR VG-CPC&7  & 6.16 &19.31 &31.41 \\
\hline
\end{tabular}
\end{center}
\caption{ABX and MAP scores (\%) for filtered representations.}
\label{tab:exp3}
\end{table} 

For tasks such as unsupervised discovery of a lexicon of word types \cite{rasanen2020unsupervised}, it is critical to accurately represent the phonemic content of word tokens without colouring by adjacent context.  Following  \cite{carlin2011rapid}, we  use the mean average precision (MAP) as defined above to assess the discriminability of \emph{word} embeddings based on the self-supervised speech representations above. We construct embeddings for each word token in the gold annotation using mean pooling. While this method is necessarily lossy, it is sufficient to demonstrate the importance of context-invariance for speech word embeddings.

For the self-supervised systems, within- and without-context ABX scores are strongly correlated. Thus, to evaluate the role of context-independence by itself, we artificially dissociate the two. We do so by applying a square filter to each of the representations: we replace each frame by the surrounding time average in a window of 3, 5, or 7 frames, pulling parts of the surrounding context into the representation of a given frame. The filter is a blurring operation to make the representations more dependent on context---in other words, less context-robust. This will have an outsized impact on the without-context condition: for within-context, the surrounding phones are identical for A, B, and X, while, for without-context, filtering will introduce substantial noise. We assess whether these  degraded representations have lower MAP scores. We perform the experiment on the clean subset again,  this time examining the across-speaker ABX as it is more relevant to the representational coherence of words (within-speaker results are qualitatively similar).

Results are shown in Table \ref{tab:exp3}. Across systems, the within-context ABX scores worsen slightly as the filter width increases (by around 0.1\% in general), while the without-context ABX scores show much greater degradation (e.g., HuBERT gets about 50\% worse in the within- condition at width 7, CPC around 12\% worse, while in the without-context condition they degrade by 90\% and 28\% respectively). The  exception is  width 3, which  has little impact.

In general, we see a relation between the without-context ABX and MAP scores. Across systems, we note much unexplained variance---the MAP scores are not entirely predictable from the ABX scores---consistent with the result of \cite{algayres2020evaluating}. Within systems, however, we see a notable degradation in the MAP scores as the (without-context) ABX scores decline. This  suggests that \emph{further} improving the context-independence of these units would lead to more consistent representations of word types, thus more useful for tasks that require accurate representation of individual word tokens.

\section{Summary of contributions}
\label{sec:summary}

This paper introduces a new version of the ZeroSpeech \textbf{ABX-LS} evaluation measure for self-supervised representations in English. The evaluation is freely available at \url{https://zerospeech.com/} and code for the additional analyses in the paper is available at  \url{https://github.com/perceptimatic/context-invariance-paper}. While the previous benchmark measures whether representations are consistent with the phonemic contrasts of the language only within specific phonetic contexts, the new benchmark measures the \emph{context-independence} of representations. We demonstrate that current systems show poor context independence: the typical case is a 300-400\% drop in performance on a context-independent (versus a context-specific) phoneme discrimination task, far larger than the gap seen for speaker-independence or for clean versus less-clean speech. We propose that future research address this gap.

\section{Acknowledgements}

\ifinterspeechfinal
     Supported by the Connaught Fund and  the Arts and Science Bridging Fund, U. of Toronto, Natural Sciences and Engineering Research Council of Canada (NSERC) RGPIN-2022-04431, ANR grants ANR-17-EURE-0017 Frontcog, ANR-10-IDEX-0001-02 PSL* and ANR-19-P3IA-0001 PRAIRIE 3IA Institute, by a Meta AI Research Gift, and by the Pronovost Morgan Family Foundation Fellowship in Ethical AI.
\else
     Supported by the ANONYMIZED ANONYMIZED ANONYMIZED  and  the ANONYMIZED ANONYMIZED ANONYMIZED ANONYMIZED, ANONYMIZED ANONYMIZED, ANONYMIZED ANONYMIZED ANONYMIZED, ANONYMIZED ANONYMIZED ANONYMIZED, ANONYMIZED ANONYMIZED ANONYMIZED  and ANONYMIZED ANONYMIZED ANONYMIZED, and by ANONYMIZED ANONYMIZED ANONYMIZED ANONYMIZED.
\fi

\bibliographystyle{IEEEtran}
\bibliography{mybib}

\end{document}